\documentclass[letterpaper, 10 pt, journal, twoside]{IEEEtran}

\usepackage{hyperref}
\usepackage[dvipsnames]{xcolor}
\usepackage{graphicx}
\usepackage{tabularx,booktabs}
\usepackage{array,multirow,graphicx}
\usepackage{makecell}
\usepackage{pdfpages}
\usepackage{amsmath}
\usepackage{amssymb}
\usepackage{mathabx}
\usepackage{booktabs} 
\usepackage{float}
\usepackage{gensymb}
\usepackage{float}
\usepackage{xcolor,colortbl}
\usepackage{soul}
\usepackage[normalem]{ulem}

\definecolor{tabfirst}{rgb}{1, 0.7, 0.7} 
\definecolor{tabsecond}{rgb}{1, 0.85, 0.7} 
\definecolor{tabthird}{rgb}{1, 1, 0.7} 

\newcommand{\ml}[2]{\multicolumn{#1}{l}{#2}}
\definecolor{LightCyan}{rgb}{0.88,0.88,0.88}
\DeclareMathOperator*{\argminA}{arg\,min} 

\newcommand{\revadd}[1]{{#1}}
\newcommand{\revdel}[1]{}

\ifCLASSINFOpdf
\else
\fi
\begin{document}
%
\title{HI-SLAM: Monocular Real-time Dense Mapping with Hybrid Implicit Fields}

%
%
%

\author{Wei~Zhang,~\IEEEmembership{Student Member,~IEEE,}
        Tiecheng~Sun,~\IEEEmembership{Member,~IEEE,}
        Sen~Wang,
        Qing~Cheng,
        Norbert~Haala%
\thanks{Manuscript received: September, 28, 2023; Accepted: December, 13, 2023. This paper was recommended for publication by Editor Sven Behnke upon evaluation of the Associate Editor and Reviewers' comments.
This work was supported by (organizations/grants which supported the work.)} 
\thanks{Wei Zhang is with the Institute for Photogrammetry, University of Stuttgart, Germany, and also with the Huawei Munich Research Center, Germany
        (email: wei.zhang@ifp.uni-stuttgart.de)}%
\thanks{Tiecheng Sun is with Central Media Technology Institute, Huawei 2012 Laboratories, China
        (email: suntiecheng1@huawei.com)}%
\thanks{Sen Wang and Qing Cheng are with the Technical University of Munich, Germany, and also with the Huawei Munich Research Center, Germany
        (email: sen.wang@tum.de; qing.cheng@tum.de)}%
\thanks{Norbert Haala is with the Institute for Photogrammetry, University of Stuttgart, Germany
        (email: norbert.haala@ifp.uni-stuttgart.de)}%
\thanks{Digital Object Identifier (DOI): see top of this page.}
}

%
%

\markboth{IEEE Robotics and Automation Letters. Preprint Version. Accepted December, 2023}
{Zhang \MakeLowercase{\textit{et al.}}: HI-SLAM: Monocular Real-time Dense Mapping with Hybrid Implicit Fields} 

%



\maketitle

\vspace*{-6mm}
\begin{abstract}

In this letter, we present a neural field-based real-time monocular mapping framework for accurate and dense Simultaneous Localization and Mapping (SLAM). Recent neural mapping frameworks show promising results, but rely on RGB-D or pose inputs, or cannot run in real-time. To address these limitations, our approach integrates dense-SLAM with neural implicit fields. Specifically, our dense SLAM approach runs parallel tracking and global optimization, while a neural field-based map is constructed incrementally based on the latest SLAM estimates. For the efficient construction of neural fields, we employ multi-resolution grid encoding and signed distance function (SDF) representation. This allows us to keep the map always up-to-date and adapt instantly to global updates via loop closing. For global consistency, we propose an efficient \textbf{$Sim(3)$}-based pose graph bundle adjustment (PGBA) approach to run online loop closing and mitigate the pose and scale drift. To enhance depth accuracy further, we incorporate learned monocular depth priors. We propose a novel joint depth and scale adjustment (JDSA) module to solve the scale ambiguity inherent in depth priors. Extensive evaluations across synthetic and real-world datasets validate that our approach outperforms existing methods in accuracy and map completeness while preserving real-time performance.

\end{abstract}

\begin{IEEEkeywords}
SLAM; Mapping; Deep Learning for Visual Perception
\end{IEEEkeywords}

%
\IEEEpeerreviewmaketitle

\vspace*{-3mm}
\section{Introduction}

\IEEEPARstart{S}{imultaneous} real-time dense mapping of the environment and camera tracking has long been a popular research topic, with vast applications in robot navigation, AR/VR, and autonomous driving. Classical SLAM approaches~\cite{klein2007parallel,qin2018vins,campos2021orb} can provide accurate camera poses, but typically yield sparse or up-to-semi-dense maps, which are insufficient for most robotic applications. Some works~\cite{tang2018ba,teed2018deepv2d} provide dense mapping, but their pose estimation is not accurate enough or they cannot generalize well to large-scale scenes. 

\begin{figure}[t!]
\centering
\vspace*{-8mm}
\includegraphics[width=0.49\textwidth]{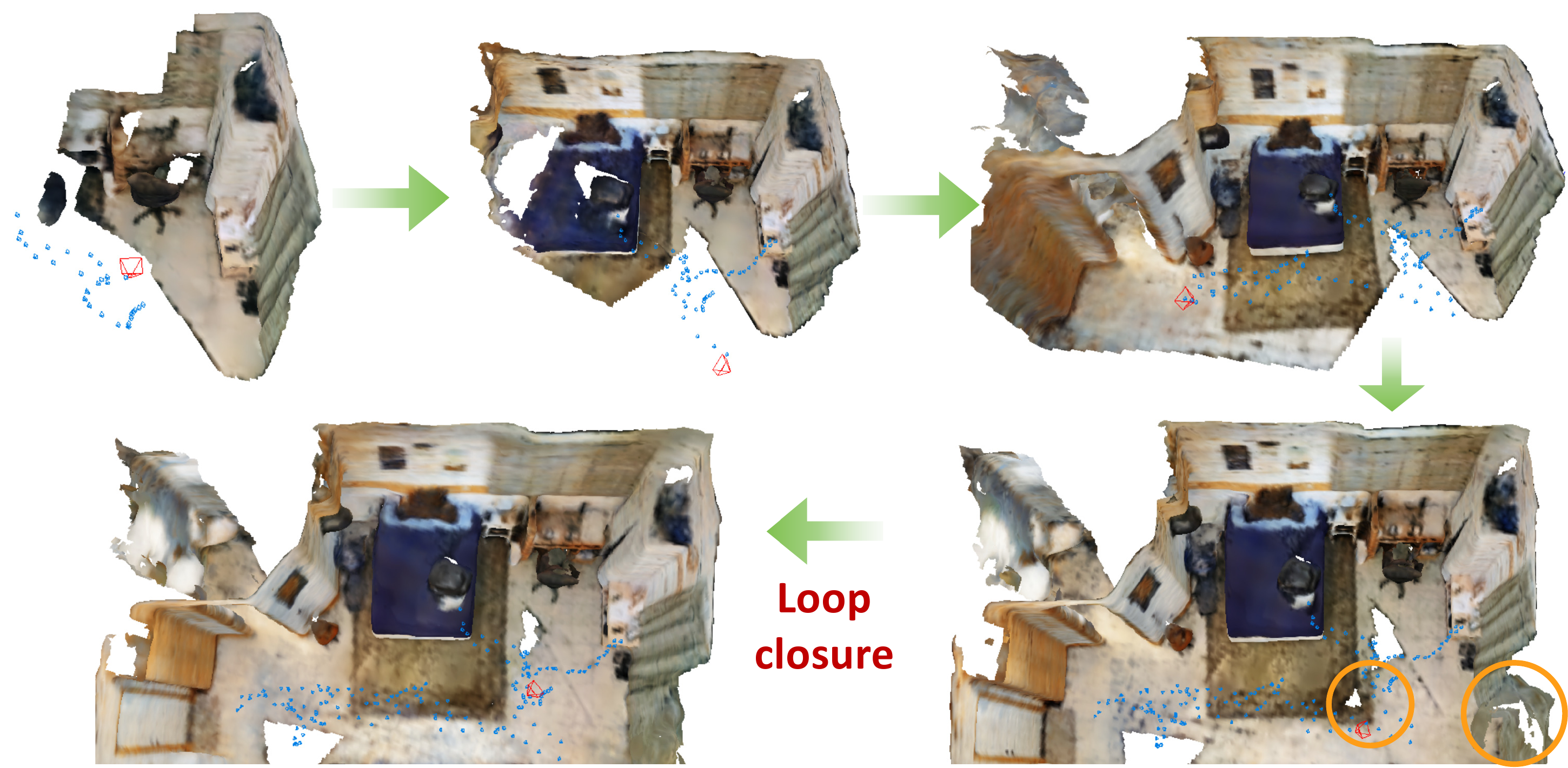}
\vspace*{-6mm}
\caption{Parallel pose tracking and dense mapping by the proposed system. In addition, our method performs on-the-fly map updates when loop closure is detected.}
\vspace*{-5mm}
\label{showcase}
\end{figure}
        
With the growth of computing resources and advances in deep learning, real-time monocular dense SLAM is now becoming feasible. Moreover, the emergence of neural implicit fields~\cite{mildenhall2020nerf} provides a new, flexible representation for dense SLAM, allowing for more complex and memory-efficient scene representation. iMAP~\cite{sucar2021imap} presents the concept of utilizing a single MLP~\cite{zhu2022nice} to jointly perform pose tracking and neural mapping. NICE-SLAM~\cite{zhu2022nice} introduces multi-resolution grids to improve efficiency. Follow-up works~\cite{yang2022vox,zhu2023nicer} apply signed distance function (SDF) for better surface definition. However, most approaches face inherent limitations, such as reliance on RGB-D~\cite{wang2023co,johari2023eslam} or fail to run in real-time~\cite{zhu2023nicer}. Moreover, previous works lack a critical aspect of SLAM: loop closure, which is essential for robots to recognize previously visited places and correct pose drift~\cite{thrun2005probabilistic}. The concurrent work~\cite{zhang2023goslam} integrates a loop closure module, but it relies on computing all-pairs co-visibility and deploying computationally intensive full BA. This considerably slows down the system and increases the risk of lost tracking. To overcome these challenges, we propose a new set-up for our SLAM system: real-time dense monocular SLAM with online loop closing and map update.

\begin{figure*}[t!]
\centering
\includegraphics[width=0.9\textwidth]{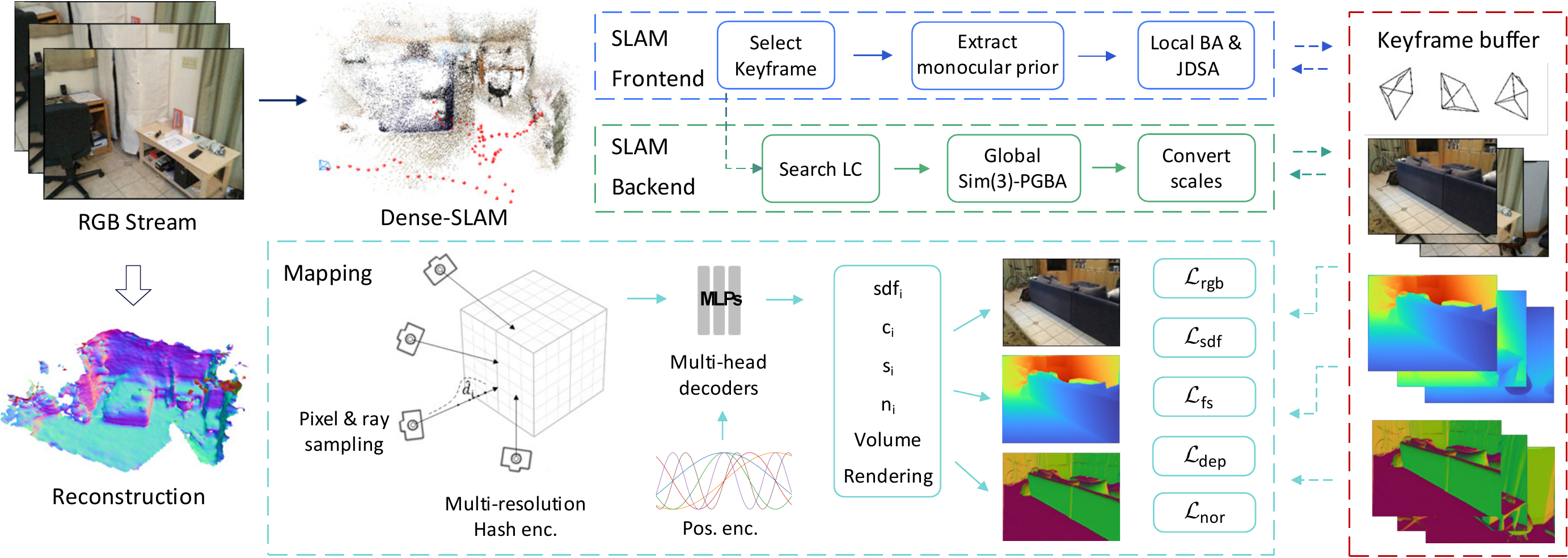}
\vspace*{-3mm}
\caption{\textbf{System overview}. Given an RGB image stream, our system runs parallel tracking and mapping. On tracking part, two processes, namely frontend and backend, are spawn for local and global consistent tracking respectively. Our SLAM frontend further leverages a pre-trained CV model to predict monocular geometric priors. The keyframe data, including estimated poses, depths, and monocular normal priors, are shared between processes. On the mapping side, the neural map is constructed incrementally based on the latest estimates from the shared buffer in an online manner.}
\label{pipeline}
\vspace*{-4mm}
\end{figure*}

Implementing such a SLAM system poses numerous challenges should be considered: real-time and dense settings demand significant computing resources. Furthermore, monocular SLAM systems face challenges including depth ambiguity, scale drift, potential slow convergence, and the risk of falling into local minima~\cite{zhu2023nicer}. Last but not least, in neural mapping systems, global updates through loop closures could be tricky to incorporate into the map as soon as possible. Delays can result in the accumulation of map artifacts due to the increasing number of processed frames and the quick collapsing of reconstruction. Therefore, it is crucial to address all of these challenges without compromising accuracy, robustness, or efficiency.

In this paper, we propose a novel approach that combines deep learning-based dense SLAM with neural implicit fields to generate dense maps in real-time without the reliance on RGB-D or pose input as in previous approaches. Our method takes full advantage of the representational capability of deep learning and the adaptability of neural implicit fields, providing a robust, efficient and accurate solution for real-time monocular dense SLAM. Furthermore, we incorporate easy-to-obtain monocular priors into our framework to recover more geometric details and maintain surface smoothness. To ensure global consistency, we run a SLAM backend in parallel to frontend tracking. This backend searches for potential loop closures and performs the proposed efficient $Sim(3)$-based pose graph bundle adjustment (PGBA). On the global update, the neural map is updated instantly according to the updated states. Fig.~\ref{showcase} shows the incremental map reconstruction process by our method, including online adaptations to loop closure updates. Through extensive experiments on both synthetic and real-world datasets, we prove that our proposed hybrid implicit dense SLAM framework can not only run in real-time but also achieve favorable accuracy and higher completeness compared to the offline methods.

The key contributions of the proposed approach are summarized as follows:
\begin{itemize}
    \item{A novel hybrid dense SLAM framework that combines the complementary features of deep learning-based dense SLAM and neural implicit fields for real-time monocular dense mapping.}
    \item{A joint depth and scale adjustment (JDSA) approach that solves the scale ambiguity of monocular depth priors and improves the quality of depth estimation.}
    \item {An efficient SLAM backend utilizing the $Sim(3)$ pose representation and pose graph BA that can correct both pose and scale drift to enable global consistent mapping.}
    \item{An effective neural map optimization scheme, which can be updated on-the-fly with rapid camera motion, and also adapt instantly with global changes by successful loop closures.}
\end{itemize}

\section{RELATED WORKS}
\textbf{Monocular dense SLAM.} Over past decades, the monocular dense SLAM technique has seen significant development. DTAM~\cite{newcombe2011dtam} pioneered one of the first real-time dense SLAM systems by parallelizing depth computing on GPU. To balance computational cost and accuracy, there are semi-dense methods~\cite{engel2014lsd}, which however do not capture texture-poor regions. In the deep learning era, many works~\cite{tang2018ba,teed2018deepv2d} stride to push the density limit by estimating dense depth maps of keyframes along with poses, but their tracking accuracy lags behind the traditional sparse landmark-based approaches. DROID-SLAM~\cite{teed2021droid} proposes to apply an optical flow network to establish dense pixel correspondences and achieve excellent trajectory estimation. Another line of works~\cite{yang2020mobile3drecon,koestler2022tandem} combines real-time VIO/SLAM systems with MVS methods for parallel tracking and dense depth estimation. The truncated signed distance function (TSDF) is then used to fuse depth maps and extract meshes. In our work, we also adopt voxel representation but store feature encodings instead of direct SDF values. This allows us to refine our map with photometric terms and regularization terms through SLAM estimations and geometric priors.

\textbf{Neural SLAM.} Recent advances in Neural Radiance Field (NeRF) have shown strong ability in scene representation. iMAP~\cite{sucar2021imap} begins to incorporate this representation into the SLAM system to perform joint neural scene and pose optimization. Later, many follow-up methods~\cite{zhu2022nice,yang2022vox,wang2023co,johari2023eslam} emerged, and neural SLAM performance was rapidly improved and reached comparable accuracy to classical methods. Nevertheless, these methods require accurate depth inputs to overcome the shortcomings of NeRF. Recently, researchers made strides to make monocular neural SLAM possible.~\cite{li2023dense,zhu2023nicer} introduce more sophisticated loss functions, such as warping loss and optical flow loss, to address the depth ambiguity problem but they cannot run in real-time and do not include loop closing. 

\textbf{Hybrid dense neural SLAM.} Recently, instead of jointly optimizing poses and neural fields, a few methods have made efforts to merge neural SLAM with classical or dense learning-based SLAM methods. Orbeez-slam~\cite{chung2023orbeez} resorts to ORB-SLAM~\cite{campos2021orb} to obtain poses and sparse point clouds, which are leveraged to regularize neural field learning. NeRF-SLAM~\cite{rosinol2022nerf} combines DROID-SLAM~\cite{teed2021droid} with InstantNGP~\cite{muller2022instant} to build real-time NeRF representation and can synthesize photo-realistic novel views.
GO-SLAM~\cite{zhang2023goslam}, concurrent to us, presents a hybrid dense SLAM system with similar spirit to ours. While it uses expensive full BA to achieve global consistency, our system can run more efficient loop closing with pose graph BA and correct scale drift. Moreover, we integrate valuable monocular priors through scale adjustment and surface regularization to boost scene geometry estimation. We achieve both higher accuracy and completeness than GO-SLAM while running faster.

\section{METHOD}
Given an RGB image stream, the goal of our framework is to simultaneously track camera poses and reconstruct high-quality and globally consistent scene geometry in real-time. Fig.~\ref{pipeline} provides an overview of our system. To achieve this, we design a multi-process pipeline to run parallel tracking and mapping. Specifically, in the tracking part, we spawn two processes to perform robust tracking (Sec.~\ref{sec:frontend}) and global optimization with loop closures (Sec.~\ref{sec:backend}). Concurrently, the mapping process reconstructs the scene incrementally using the continuously updated states estimated by the SLAM frontend and backend processes (Sec.~\ref{sec:map}).

\subsection{Robust Frontend Tracking}\label{sec:frontend}
\begin{figure}[t!]
\centering
\includegraphics[width=0.49\textwidth]{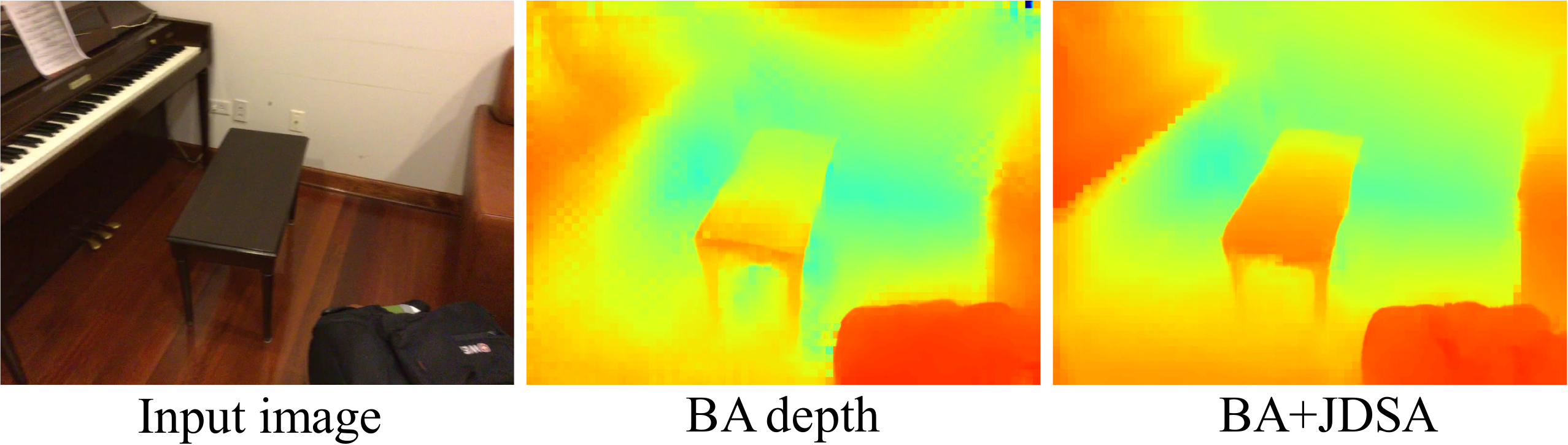}
\vspace*{-7mm}
\caption{Comparison of depths with and without incorporating depth prior by JDSA module.}
\label{compdepth}
\vspace*{-4mm}
\end{figure}

To robustly track the camera poses under challenging scenarios, such as low texture and rapid movement, we build our SLAM system on the foundation of DROID-SLAM~\cite{teed2021droid}, which adopts optical flow network to accurately predict dense pixel correspondences between nearby frames. Our system maintains a keyframe graph $(\mathcal{V}, \mathcal{E})$ representing the co-visibility of keyframes and a keyframe buffer storing keyframe information and their respective states. For each incoming frame, the mean flow distance to the last keyframe is determined by a single-pass through the optical flow network. If the distance surpasses a predefined threshold $d_{flow}$, the current frame is selected as a keyframe and added to the buffer. The edges between this new keyframe and its neighbors are inserted into the keyframe graph. Besides, nearby keyframes with high co-visibility, namely those with small flow distance, are also linked to the latest keyframe. These edges extend the tracking duration of each view. Local BA is then performed based on a co-visibility graph within a sliding window. We employ the flow predictions by the network as targets, denoted as $\mathbf{\check{p}}_{ij}$, and refine the poses $\mathbf{T}$ and depth maps $\mathbf{d}$ of the keyframes involved. This BA optimization is solved iteratively using a damped Gauss-Newton algorithm with the following objective:
\begin{equation}\label{equ:1}
\argminA_{\mathbf{T},\mathbf{d}} \sum_{(i,j) \in \mathcal{E}}^{} \| \mathbf{\check{p}}_{ij} - \Pi(\mathbf{T}_{ij} \Pi^{-1}(\mathbf{p}_i, \mathbf{d}_i)) \|_{\Sigma_{ij}}^2
\end{equation}
where \revadd{$\mathbf{d}_i$ refers to the depth map of keyframe $i$ in inverse depth parametrization, $\Pi$ and $\Pi^{-1}$ are the projection and back-projection functions}, $\Sigma_{ij}$ is a diagonal matrix composed of the prediction confidences by the network. This matrix serves to weigh down the influence of occluded and hard-to-match pixels. However, in this way, the pixels with low confidences attain high depth variances in occluded or texture-poor regions and can not achieve accurate depth estimation, as depicted in Fig.~\ref{compdepth}. To address this, we extend the system by incorporating monocular depth priors.

\textbf{Incorporate monocular depth prior.}
The depth map estimation is largely dependent on the matching ability of the network, which is highly based on image texture. In regions with low-texture, the network often struggles to find robust correspondences. As shown in Fig.~\ref{fig:depth}, depth certainties (inverse of variances) are low on texture poor regions such as white tables and walls. To this end, we leverage the off-the-shelf monocular depth network~\cite{eftekhar2021omnidata} to generate depth priors. This can then be incorporated into our BA optimization process. Although this network is versatile and can generalize to unseen scenes, the predicted depths prior are relative with varying scales across different viewpoints.

To address this, for each depth prior \revadd{$\check{\mathbf{d}}_i$}, we estimate a scale $s_i$ and \revdel{and} offset $o_i$. We then attempt to incorporate them as variables within the BA optimization. The depth prior factor can be formulated as:
\begin{equation}\label{equ:dprior}
	r_{d} = \|( \check{\mathbf{d}}_i \cdot s_i + o_i ) - \mathbf{d_i} \|^2 
\end{equation}
\revdel{where $\check{d}_i$ and $\hat{d_i}$ denote the prior depth and the estimated depth respectively in inverse depth representation.} We observe that the prior scales may not necessarily converge when jointly optimized with camera poses. The scales of camera poses could be misdirected by scale-varying priors to shift away. Thus, we explore alternative methods to separate the monocular scale estimation from the BA problem. Specifically, we introduce a joint depth and scale adjustment (JDSA) module. This module minimizes both reprojection factors and depth prior factors, while keeping camera poses fixed.
\revadd{Fixing the poses in this JDSA module prevents scale drift which can arise from the scale-varying priors, and the prior scales can be properly aligned to the system scale.}
\revdel{This approach ensures that camera poses remain unaffected by the ambiguous scales of the monocular priors.} The objective of the JDSA problem can be formulated as:
\begin{equation}\label{equ:JDSA}
    \begin{aligned}
	\argminA_{\mathbf{d}, \mathbf{s,o}} &\sum_{(i,j) \in \mathcal{E}}^{} \| \check{\mathbf{p}}_{ij} - \Pi(\mathbf{T}_{ij} \Pi^{-1}(\mathbf{p}_i, \mathbf{d}_i)) \|_{\Sigma_{ij}}^2 + \\
     &\ \ \sum_{i \in \mathcal{V}}^{} \ \ \|( \check{\mathbf{d}}_i \cdot s_i + o_i ) - \mathbf{d}_i \|^2
    \end{aligned}
\end{equation}
In each iteration, we alternate between the BA and JDSA optimizations, which can complement to each other. The poses estimated by BA ensure the consistent scales of depth priors in JDSA optimization. In return, the JDSA provides refined depths that facilitate easier flow updates. This updates are then converted into poses and depths via BA. Fig.~\ref{compdepth} shows the enhanced depth accuracy, especially in low texture areas such as white wall and black bench.

\subsection{Backend with Global Consistent Optimization}\label{sec:backend}
While the SLAM frontend can reliably track accurate camera poses, pose drift can accumulate inevitably over long distances. Finding loop closing and performing global optimization is an effective way to minimize this drift error. Moreover, due to inherent scale ambiguity, monocular SLAM methods also face scale drift. We first introduce how we detect loop closures and then present our proposed $Sim(3)$-based pose graph BA, designed for efficient loop closing in an online system.

\textbf{Loop closure detection.}
Loop closure detection runs in parallel to the tracking process. For each new keyframe, we compute the flow distances $d_{of}$ between the new keyframe and previous keyframes. Three criteria are defined for selecting loop closure candidates. Firstly, $d_{of}$ should fall below a predefined threshold $\tau_{flow}$, ensuring adequate co-visibility for successful convergence of recurrent flow updates. Secondly, orientation differences based on current pose estimation should remain below a threshold $\tau_{ori}$. Lastly, the difference in frame indices should be at least $\tau_{temp}$ indices. If all criteria are satisfied, we add edges between the selected keyframe pairs bidirectionally into our keyframe graph.

\textbf{$Sim(3)$-based pose graph BA.}
Upon a few loop closure candidates are detected, inspired by~\cite{wei2023bamf}, we opt for pose graph BA over full BA to enhance efficiency while maintaining accuracy. To tackle scale drift, we use $Sim(3)$ to represent keyframe poses allowing for scale updates. Before each run, we convert the latest pose estimates from $SE(3)$ to $Sim(3)$ and initialize the scales with ones. The pixel warping step follows Eq.~\ref{equ:1}, but the $SE(3)$ transformation is replaced by $Sim(3)$ transformation.

Another aspect of pose graph BA is to construct a pose graph connected by relative pose edges. Follows~\cite{wei2023bamf}, we compute relative poses from the dense correspondences of inactive reprojection edges. These come into play when their associated keyframes exit the sliding window of the frontend. Having been refined multiple times while active in the sliding window, these dense correspondences offer a solid foundation for computing relative poses. The same reprojection error term in Eq.~\ref{equ:1} is used but only optimizing for the relative poses $\mathbf{\check{T}}_{ij}$ under the assumption the depths are already accurately estimated. Along with the estimated relative poses, the associated variances $\Sigma^{rel}_{ij}$ are estimated based on the adjustment theory~\cite{niemeier2008ausgleichungsrechnung} as:
\begin{equation}
    \Sigma^{rel}_{ij} = (\mathbf{J} \Delta \mathbf{T_{ij}} - \mathbf{r})^T \Sigma_{ij} (\mathbf{J} \Delta \mathbf{T_{ij}} - \mathbf{r}) (\mathbf{J}^T \Sigma_{ij} \mathbf{J})^{-1}
\end{equation}
where $\mathbf{J}$, $\mathbf{r}$ and $\Delta \mathbf{T_{ij}}$ are the Jacobian, the reprojection residuals, and the relative pose update from the last iteration. The relative pose variances serve as weights for pose graph BA. Finally, the objective PGBA problem is to minimize the sum of the relative pose factors and reprojection factors as:
\begin{equation}\label{equ:pgba}
    \begin{aligned}
	\argminA_{\mathbf{T}, \mathbf{d}} &\sum_{(i,j) \in \mathcal{E}^*}^{} \| \mathbf{\check{p}}_{ij} - \Pi(\mathbf{T}_{ij} \Pi^{-1}(\mathbf{p}_i, \mathbf{d}_i)) \|_{\Sigma_{ij}}^2 + \\
     & \sum_{(i,j) \in \mathcal{E}^{+}}^{} \| \log(\mathbf{\check{T}}_{ij} \cdot \mathbf{T}_{i} \cdot \mathbf{T}_{j}^{-1}) \|_{\Sigma^{rel}_{ij}}^2
    \end{aligned}
\end{equation}
where $\mathcal{E}^*$ and $\mathcal{E}^+$ are the set of detected loop closures and the set of relative pose factors respectively.

\subsection{Hybrid Scene Representation}\label{sec:map}
For our scene representation, we first leverage multi-resolution hash feature grid~\cite{muller2022instant} denoted as $h_{\theta_{hash}}$ with optimizable parameters $\theta_{hash}$. For a sample point $\mathbf{x}$, features at every resolution are looked up via tri-linear interpolation and concatenated together. This yields a coarse-to-fine feature encoding. To enhance the scene completion capability, inspired by~\cite{mildenhall2020nerf,wang2023co}, we also adopt the positional encoding $\gamma(\mathbf{x})$ to map the coordinate to a encoding in high-dimensional spaces. Both the hash and positional encodings serve both spatial and geometric purposes and are fed into our SDF network.

Our SDF network, denoted as $f_{\theta_{sdf}}$, serves as a geometry decoder. It predicts a SDF value $s$ and a geometric feature vector $h$ expressed as:
\begin{equation}\label{equ:2}
    s, \mathbf{h} = f_{\theta_{sdf}}(\gamma(\mathbf{x}), h_{\theta_{hash}}(\mathbf{x}))
\end{equation}
where $\theta_{sdf}$ represents the learnable parameters of our SDF network, which is a shallow MLP with two 32-dimension hidden layers. 
Following~\cite{yu2022monosdf} and given the differentiability of both the hash feature grid and SDF network, we can compute the analytical gradient of the SDF function $\nabla f_{\theta_{sdf}} $. After normalization, this gradient becomes the surface normal $n$.

Similar to our SDF decoder, we also employ a color network to predict the color value $c$ as:
\begin{equation}\label{equ:3}
    \mathbf{c} = f_{\theta_{color}}(\gamma(\mathbf{x}), \mathbf{h})
\end{equation}
where $\theta_{color}$ denotes the learnable parameters of the color network, which shares the same MLP architecture as our SDF network.

\begin{figure}[t!]
\centering
\includegraphics[width=0.44\textwidth]{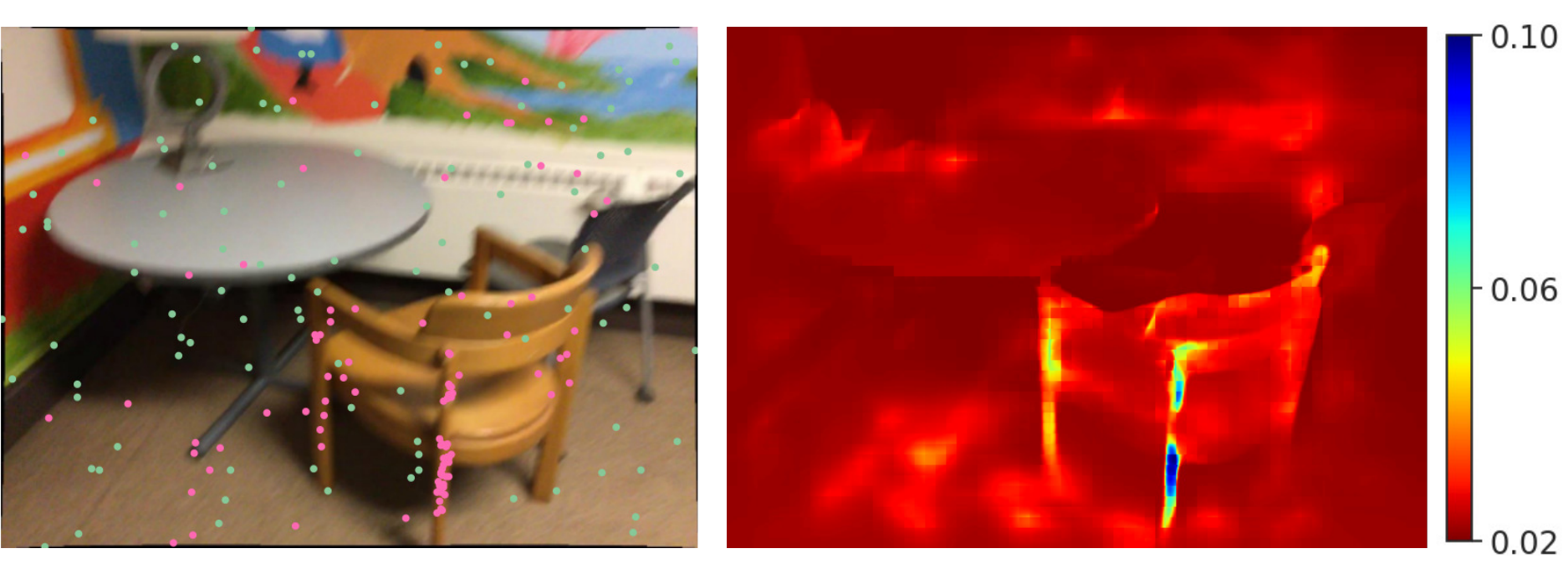}
\vspace*{-4mm}
\caption{Left: example sample pixels of \textcolor{CarnationPink}{certainty-guided}, \textcolor{YellowGreen}{uniform-based}; Right: depth certainties.}
\label{fig:depth}
\vspace*{-6mm}
\end{figure}

\textbf{Depth-guided pixel and ray sampling.} From the keyframe buffer, the latest depth estimations with associated variances provides readily the guidance of neural map optimization. During each map optimization step, we sample $N_{pixel}$ pixels from the current keyframe buffer. While NeRF and many follow-up methods use straightforward uniform pixel sampling, we observe that they tend to poorly reconstruct small objects or smooth out boundaries, due to insufficient sampling in these areas. To address this, we propose a depth certainty-guided importance sampling strategy. Specifically, we sample $N_{pixel} / 2$ pixels based on the depth variance of each pixel. Pixels with lower variance (means higher certainty) are sampled more frequently. As illustrated in Fig.~\ref{fig:depth}, pixels on object borders typically have higher certainties than those on the flat surfaces. For the remaining $N_{pixel} / 2$ pixels, we combine uniform sampling to ensure coverage on low-texture areas, such as floors and walls.

After the pixel sampling step, rays are cast from the optical center through the sampled pixels, and we sample query points along these rays. Inspired by~\cite{wang2023co}, we first sample $M_{d}$ points centered around the estimated depth. In addition, $M_{u}$ points are uniformly sampled between the predefined near and far bounds. For each ray, we sample a total of $M=M_u + M_d$ query points.

Following the bell-shaped formulation in~\cite{azinovic2022neural}, we can re-render the depth and color value of a pixel. We first compute the weights of each sample point from its predicted signed distance values as:
\begin{equation}\label{equ:4}
    w_{i} = \sigma(\frac{s_i}{tr})\cdot \sigma(-\frac{s_i}{tr})
\end{equation}
where $tr=10\,cm$ denotes the truncation distance. We then normalize the weights by the sum of all values on each ray. Using this weights, we can calculate the rendered depth, color, and normal by accumulating the values along the ray~\cite{yariv2021volume} as:
\begin{equation}\label{equ:5}
    \mathbf{\hat{C}} = \sum_{i = 1}^{M}w_i \mathbf{c_i},  \ \ \ 
    \hat{D} = \sum_{i = 1}^{M}w_i d_i, \ \ \
    \mathbf{\hat{N}} = \sum_{i = 1}^{M}w_i \mathbf{n_i}
\end{equation}

\begin{figure}[t!]
\hspace*{-3mm}
\includegraphics[width=0.496\textwidth]{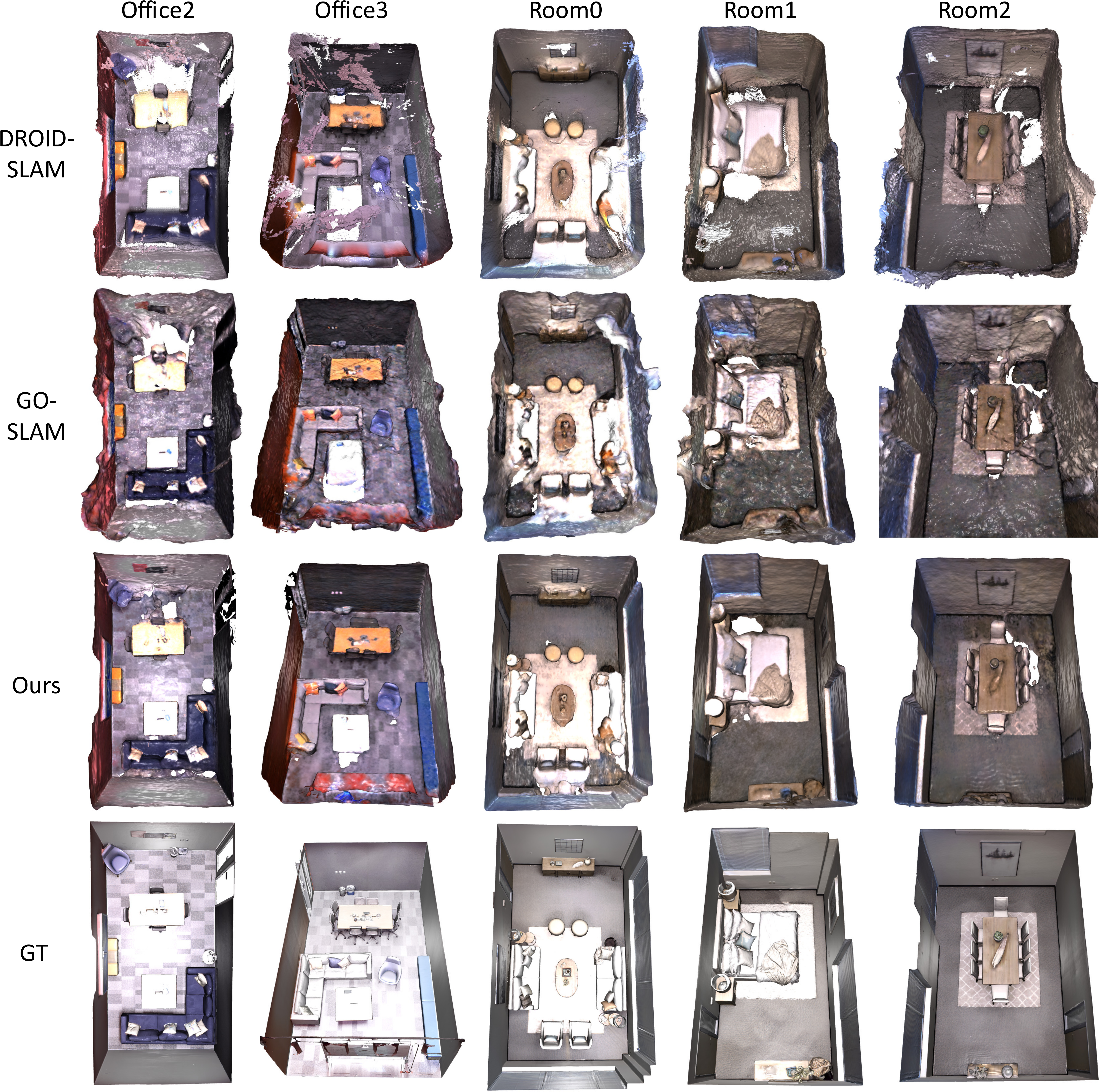}
\vspace*{-6mm}
\caption{Reconstruction results on Replica dataset~\cite{straub2019replica}. Our results show smoother surfaces with more detailed geometric compared to DROID-SLAM~\cite{teed2021droid} and GO-SLAM~\cite{zhang2023goslam}.}
\label{replica}
\vspace*{-6mm}
\end{figure}

\textbf{Optimization losses.}
Our neural map optimization is performed based on various objective functions with respect to the learnable parameters $\theta = \{\theta_{hash}, \theta_{sdf}, \theta_{color}\}$. Following~\cite{mildenhall2020nerf}, we first apply the color rendering loss, which computes errors between the rendered colors and input image colors as:
\begin{equation}\label{equ:6}
    \mathcal{L} _{c} = \frac{1}{N} \sum_{n=1}^{N} (\mathbf{\hat{C}_n} - \mathbf{C}_n) ^ 2
\end{equation}
Likewise, the rendered depths are supervised by estimated depths from dense SLAM:
\begin{equation}\label{equ:7}
    \mathcal{L}_{d} = \frac{1}{N} \sum_{n=1}^{N} \frac{(\hat{D}_n - D_n)^2}{\sigma_n ^ 2}
\end{equation}
where $\sigma_n ^ 2$ is the associated depth variance derived from the Hessian matrix of the BA problem~\cite{rosinol2023probabilistic}. This effectively reduces the influence of uncertain noisy depths. Furthermore, we supervise the surface normal prediction by the monocular normal prior $\mathbf{N}_n$ by pre-trained Omnidata network~\cite{eftekhar2021omnidata} as:
\begin{equation}\label{equ:normal}
    \mathcal{L}_{n} = \frac{1}{N} \sum_{n=1}^{N}(\mathbf{\hat{N}}_n - \mathbf{N}_n)
\end{equation}
where $\mathbf{N}_n$ is transformed from the local coordinate frame to the global one using the currently estimated poses. This ensures consistency with the SDF gradient that is in the global frame.

To accelerate training, following~\cite{yang2022vox,wang2023co}, we also directly supervise the SDF predictions. For those points within the truncation bounds, namely the point set $S^{tr}$ where $\vert D - d_i \vert \le tr$, we ensure that the neural fields learn to approximate a surface distribution. This is achieved using the pseudo-ground-truth SDF values derived from the estimated depths:
\begin{equation}\label{equ:8}
    \mathcal{L}_{sdf} = \frac{1}{N} \sum_{n=1}^{N} \frac{1}{|S^{tr}|} \sum_{p\in S^{tr}} (s_p - (D-d_i) )^2
\end{equation}
For the sampled points outside the truncation bounds, denoted as $S^{fs}$, we enforce the SDF prediction to match the truncation distance $tr$ in order to encourage the prediction in free space:
\begin{equation}\label{equ:9}
    \mathcal{L}_{fs} = \frac{1}{N} \sum_{n=1}^{N} \frac{1}{|S^{fs}|} \sum_{p\in S^{fs}} (s_p - tr )^2
\end{equation}
Finally, our total loss can be formulated as:
\begin{equation}\label{equ:total}
    \mathcal{L} = \lambda_c \mathcal{L}_c + \lambda_d \mathcal{L}_d + \lambda_n \mathcal{L}_n + \lambda_{sdf} \mathcal{L}_{sdf} + \lambda_{fs} \mathcal{L}_{fs}
\end{equation}
where we assign the loss weights $\lambda_c,\lambda_d,\lambda_n,\lambda_{sdf}$ and $\lambda_{fs}$ to 10, 0.1, 1, 1000, and 2 respectively. This ensures a balance between the photometric and various geometric supervisions. For each newly created keyframe, we optimize our neural map representation for 10 iterations. For global updates via loop closures, we extend the optimization process to 50 iterations accounting for greater changes.
\revadd{While this is generally effective for major map changes, it might fall short in refining the finer details. However, subsequent map updates on new keyframes can continually enhance the details of the updated regions, progressively improving the overall scene quality}.

\begin{figure*}[t!]
\centering
\includegraphics[width=0.996\textwidth]{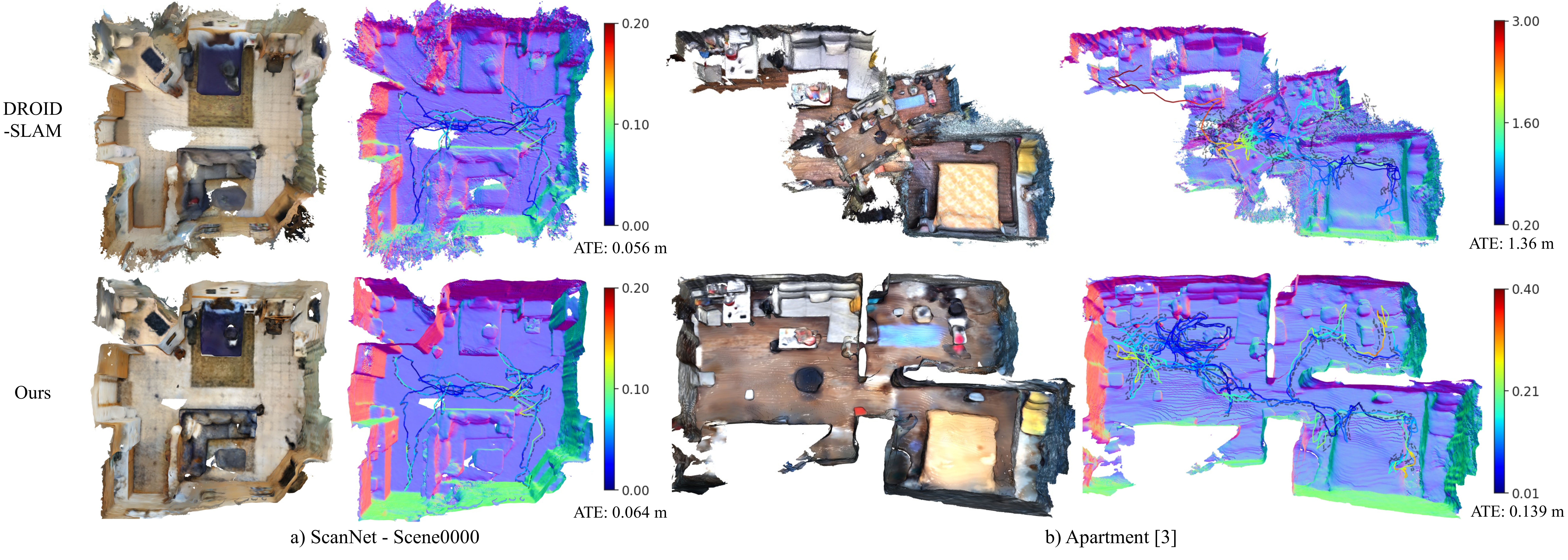}
\vspace*{-9mm}
\caption{Qualitative results of reconstructed maps and estimated trajectories colored by ATE. Our method can produce more complete maps while using less memory footprint, object surfaces are smoother with finer details. On the apartment dataset, DROID-SLAM fails to detect all loop closures and suffers from severe scale drift. Please check our attached demo videos\textsuperscript{1} for the incremental mapping process.}
\label{quality-scene0000}
\vspace*{-4mm}
\end{figure*}

\section{EXPERIMENTS}
We evaluate our proposed system on both synthetic and real-world datasets, including Replica~\cite{straub2019replica}, ScanNet~\cite{dai2017scannet}, and the larger-scale Apartment dataset from NICE-SLAM~\cite{zhu2022nice}. Both tracking accuracy and reconstruction quality metrics are reported and compared with prior works. Additionally, we conduct an ablation study to validate the effectiveness of the proposed components. Finally, a runtime analysis is provided.

\setlength{\tabcolsep}{3pt}
\begin{table}[h]
\centering
\caption{Reconstruction evaluations on Replica dataset. \revdel{Upper part is the evaluation based on complete GT mesh to consider the scene completion ability, while lower part focuses on evaluating the observed region only.}\revadd{Best results are highlighted as \colorbox{tabfirst}{first}, \colorbox{tabsecond}{second}, and \colorbox{tabthird}{third}}}\label{tab:replica}
\vspace{-3mm}
\resizebox{0.495\textwidth}{!}{
\begin{tabular}{ccccccccccc}
\toprule
& & ro-0 & ro-1 & ro-2 & of-0 & of-1 & of-2 & of-3 & of-4 & Avg.\\
\midrule
\multirow{3}{*}{\rotatebox[origin=c]{90}{iMAP}} & \ml{1}{Acc. [cm]$\downarrow$} & \cellcolor{tabthird}3.58      &       \cellcolor{tabsecond}3.69       &       4.68    &       5.87    &       3.71    &       4.81    &        \cellcolor{tabthird}4.27       &       4.83    &  4.43\\
& \ml{1}{Comp. [cm]$\downarrow$} & \cellcolor{tabthird}5.06     &       4.87    &        \cellcolor{tabthird}5.51       &       6.11    &        \cellcolor{tabthird}5.26       &       5.65    &       5.45    &       6.59    &       5.56\\
& Comp. Ratio [\%] & \cellcolor{tabthird}83.90  &       \cellcolor{tabsecond}83.40      &        \cellcolor{tabthird}75.50      &       \cellcolor{tabsecond}77.70      &       \cellcolor{tabsecond}79.60      &       \cellcolor{tabsecond}77.20      &        \cellcolor{tabthird}77.30  &        \cellcolor{tabthird}77.60      &        \cellcolor{tabthird}79.00\\
\midrule
\multirow{3}{*}{\rotatebox[origin=c]{90}{iMODE}} & \ml{1}{Acc. [cm]$\downarrow$} &7.40  &       6.40    &       9.30    &       6.60    &       11.80   &       11.40   &       9.40    &       8.00    &       8.78\\
& \ml{1}{Comp. [cm]$\downarrow$} &13.50 &       10.10   &       19.20   &       9.70    &       17.00   &       14.50   &       11.80   &       15.40   &       13.90\\
& Comp. Ratio [\%] &38.70       &       46.10   &       36.10   &       49.30   &       30.10   &       29.80   &       36.00   &       31.00   &       37.10\\
\midrule
\multirow{3}{*}{\rotatebox[origin=c]{90}{DROID}} & \ml{1}{Acc. [cm]$\downarrow$} &12.18 &       8.35    &       \cellcolor{tabsecond}3.26       &        \cellcolor{tabfirst}3.01       &        \cellcolor{tabfirst}2.39       &       5.66    &       4.49    &       4.65    &  5.50\\
& \ml{1}{Comp. [cm]$\downarrow$} &8.96  &       6.07    &       16.01   &       16.19   &       16.20   &       15.56   &       9.73    &       9.63    &       12.29\\
& Comp. Ratio [\%] &60.07       &       76.20   &       61.62   &       64.19   &       60.63   &       56.78   &       61.95   &       67.51   &       63.60\\
\midrule
\multirow{3}{*}{\rotatebox[origin=c]{90}{NICER}} & \ml{1}{Acc. [cm]$\downarrow$} & \cellcolor{tabfirst}2.53     &       3.93    &        \cellcolor{tabthird}3.40       &       5.49    &       3.45    &        \cellcolor{tabfirst}4.02       &        \cellcolor{tabfirst}3.34  & \cellcolor{tabfirst}3.03       &       \cellcolor{tabsecond}3.65\\
& \ml{1}{Comp. [cm]$\downarrow$} & \cellcolor{tabfirst}3.04     &        \cellcolor{tabthird}4.10       &        \cellcolor{tabfirst}3.42       &        \cellcolor{tabthird}6.09       &       \cellcolor{tabsecond}4.42       &        \cellcolor{tabfirst}4.29       &        \cellcolor{tabfirst}4.03   &       \cellcolor{tabsecond}3.87       &        \cellcolor{tabfirst}4.16\\
& Comp. Ratio [\%] & \cellcolor{tabfirst}88.75  &       76.61   &        \cellcolor{tabfirst}86.10      &       65.19   &        \cellcolor{tabthird}77.84      &       74.51   &        \cellcolor{tabfirst}82.01      &        \cellcolor{tabfirst}83.98      &       \cellcolor{tabsecond}79.37\\
\midrule
\multirow{4}{*}{\rotatebox[origin=c]{90}{GO-SLAM}} & \ml{1}{Depth L1 [cm]$\downarrow$} & -&-&-&-&-&-&-&-&\cellcolor{tabsecond}4.39\\
& \ml{1}{Acc. [cm]$\downarrow$} &4.60        &        \cellcolor{tabfirst}3.31       &       3.97    &       \cellcolor{tabsecond}3.05       &        \cellcolor{tabthird}2.74       &       \cellcolor{tabsecond}4.61       &  4.32     &        \cellcolor{tabthird}3.91       &        \cellcolor{tabthird}3.81\\
& \ml{1}{Comp. [cm]$\downarrow$} &5.56  &       \cellcolor{tabsecond}3.48       &       6.90    &        \cellcolor{tabfirst}3.31       &        \cellcolor{tabfirst}3.46       &        \cellcolor{tabthird}5.16       &        \cellcolor{tabthird}5.40       &        \cellcolor{tabthird}5.01   &        \cellcolor{tabthird}4.79\\
& Comp. Ratio [\%] &73.35       &        \cellcolor{tabthird}82.86      &       74.23   &        \cellcolor{tabfirst}82.56      &        \cellcolor{tabfirst}86.19      &        \cellcolor{tabthird}75.76      &       72.63   &       76.61   &       78.00\\
\midrule
\multirow{4}{*}{\rotatebox[origin=c]{90}{Ours}} & \ml{1}{Depth L1 [cm]$\downarrow$} & 3.61& 2.07& 4.63& 3.66& 1.91& 3.39& 5.07& 4.68& \cellcolor{tabfirst}3.63\\
& \ml{1}{Acc. [cm]$\downarrow$} &\cellcolor{tabsecond}3.21      &        \cellcolor{tabthird}3.74       &        \cellcolor{tabfirst}3.16       &        \cellcolor{tabthird}3.87       &       \cellcolor{tabsecond}2.60       &   \cellcolor{tabthird}4.62        &       \cellcolor{tabsecond}4.25       &       \cellcolor{tabsecond}3.53       &        \cellcolor{tabfirst}3.62\\
& \ml{1}{Comp. [cm]$\downarrow$} &\cellcolor{tabsecond}3.25     &        \cellcolor{tabfirst}3.08       &       \cellcolor{tabsecond}4.09       &       \cellcolor{tabsecond}5.29       &       8.83    &       \cellcolor{tabsecond}4.42       &       \cellcolor{tabsecond}4.06  & \cellcolor{tabfirst}3.72       &       \cellcolor{tabsecond}4.59\\
& Comp. Ratio [\%] &\cellcolor{tabsecond}86.99  &        \cellcolor{tabfirst}87.19      &       \cellcolor{tabsecond}80.82      &        \cellcolor{tabthird}72.55      &       72.44   &        \cellcolor{tabfirst}80.90      &       \cellcolor{tabsecond}81.04      &       \cellcolor{tabsecond}82.88  &        \cellcolor{tabfirst}80.60\\
\bottomrule
\vspace{-3mm}
\end{tabular}}
\vspace{-4mm}
\end{table}

\subsection{Implementation Details}
We build our BA and neural map optimization using the PyTorch library and employ LieTorch library~\cite{teed2021tangent} for pose representation in $SE(3)$ and $Sim(3)$ groups. We use the pre-trained model from DROID-SLAM~\cite{teed2021droid} and input images are resized to 400x536 to better align with the resolution of training images and enhance efficiency. \revadd{For multiresolution features grids, we utilize the tiny-cuda-nn framework, which implements the fast fully-fused CUDA kernels to accelerate computing}. We set 16 grid levels ranging from a base resolution 16 to a finest spacing of 4\,cm. The hash table size is set to $2^{16}$ for all room-size datasets and increased to $2^{19}$ for Apartment data considering its larger dimensions. For map optimization, we sample a total of $N_{pixel}=2048$ pixels in each iteration. Along each sampled pixel ray, we sample $M_d=11$ plus $M_u=32$ query points.
\footnotetext[1]{https://youtu.be/lj4Ie1RBFBE?si=zB7XWqwS6egdEexL}

\setlength{\tabcolsep}{5pt}
\begin{table}[h]
\centering
\caption{ATE [m] results on ScanNet dataset.}\label{tab:scannet}
\vspace{-3mm}
\resizebox{0.48\textwidth}{!}{
\begin{tabular}{ccccccccc}
\toprule
& \ml{1}{Scene ID} & 0000 & 0059 & 0106 & 0169 & 0181 & 0207 & Avg.\\
\midrule
\multirow{3}{*}{\rotatebox[origin=c]{90}{RGB-D}} & \ml{1}{NICE-SLAM\,\cite{zhu2022nice}} & 0.086& 0.123& 0.081& 0.103& 0.129& \textbf{0.056} & 0.096\\
& \ml{1}{Co-SLAM\,\cite{wang2023co}} & 0.072& 0.123& 0.096& 0.066& 0.134& 0.071& 0.094\\
& \ml{1}{ESLAM\,\cite{johari2023eslam}} & 0.073& 0.086& 0.075& \textbf{0.065}& 0.092& 0.057& \textbf{0.074}\\
\midrule
\multirow{5}{*}{\rotatebox[origin=c]{90}{Mono.}} & \ml{1}{GO-SLAM\,\cite{zhang2023goslam}} &0.059& 0.083& 0.081& 0.084& 0.083 & - & - \\
& \ml{1}{DROID-SLAM\,\cite{teed2021droid}\,(VO)} & 0.145 & 0.281& 0.088& 0.180& 0.089& 0.102& 0.148\\
& \ml{1}{DROID-SLAM\,\cite{teed2021droid}} & \textbf{0.056}& 0.080& 0.066& 0.092& 0.077& 0.076& \textbf{0.074}\\
& \ml{1}{Ours\,(VO)} & 0.144 & 0.267& 0.100& 0.155& 0.093& 0.099& 0.143\\
& \ml{1}{Ours} & 0.064& \textbf{0.072} & \textbf{0.065} & 0.085& \textbf{0.076} & 0.084& \textbf{0.074}\\
\bottomrule
\vspace{-7mm}
\end{tabular}}
\end{table}

\subsection{Evaluation on Replica dataset}
Replica~\cite{straub2019replica} comprises several high-quality reconstructions from real-world scenes. We use the synthesized sequences by~\cite{sucar2021imap} as the benchmark for comparison against prior works. We omit the trajectory evaluation for this dataset since both our method and previous works have already achieved excellent accuracy below 1\,cm. \revdel{For mapping evaluation, two evaluation protocols are commonly used. NICE-SLAM~\cite{zhu2022nice} cull the ground-truth (GT) model by the observability of input frames so only the observed areas are evaluated, while iMAP takes the complete GT model as the reference to test the scene completion capability of the neural implicit map. For comparison with the culled GT mesh, we also culled our reconstructions by the visibility of input frames as done by prior works~\cite{zhu2023nicer}~\cite{zhang2023goslam}.}\revadd{Notably, neural field-based map has the scene completion capability. In alignment with \cite{sucar2021imap} \cite{zhu2023nicer}, we take the complete ground-truth (GT) models to also assess the reconstruction quality on unseen areas. Additionally, following the approach of \cite{zhu2022nice}, a convex hull is computed based on the keyframe poses and rendered depth maps, and mesh faces outside this are considered outliers.}

As shown in Tab.~\ref{tab:replica}, our results exhibit superior performance in both accuracy and completeness. Notably, \revdel{when compared to the complete GT mesh} our results are comparable to the RGB-D approach iMAP~\cite{sucar2021imap} and significantly outperform the \revdel{prior} \revadd{previous neural field-based} RGB method iMODE~\cite{matsuki2023imode}. \revdel{When using the culled GT mesh for comparison, DROID-SLAM is included as a baseline with its mesh extracted from the TSDF fusion of its predicted depths} \revadd{We also compare with the dense-SLAM approach, specifically DROID-SLAM, where reconstruction is based on the TSDF integration of predicted depths}. Furthermore, NICER-SLAM~\cite{zhu2023nicer} and GO-SLAM~\cite{zhang2023goslam}, \revadd{recent state-of-the-art methods}, serve as another two strong baselines. While NICER-SLAM also employs monocular priors to facilitate neural scene reconstruction, it is unable to run in real-time nor achieve loop closure optimization. Fig.~\ref{replica} shows qualitatively that our reconstruction are cleaner with more detailed geometry.

\setlength{\tabcolsep}{3pt}
\begin{table}[h]
\centering
\caption{Quantitative evaluation of the reconstruction on ScanNet. Averaged on 4 selected sequences. We also report the runtime on scene0059 by each method in last column.}\label{tab:scannetrecon}
\vspace{-3mm}
\resizebox{0.49\textwidth}{!}{
\begin{tabular}{cccccccc}
\toprule
\ml{1}{Method} & Pose & Acc$\downarrow$ & Comp$\downarrow$ & Prec$\uparrow$ & Recall$\uparrow$ & F-score$\uparrow$ & Time[h]$\downarrow$\\
\midrule
\ml{1}{ManhattanSDF\,\cite{guo2022neural}} & GT & 0.072& 0.068& 0.621& 0.586& 0.602 & 16.68\\
\ml{1}{MonoSDF\,\cite{yu2022monosdf}\,(MLP)} & GT & \textbf{0.031}& 0.057& 0.783& 0.652& 0.710 & 9.89\\
\ml{1}{MonoSDF\,\cite{yu2022monosdf}\,(Grid)} & GT & 0.034& 0.046& \textbf{0.796}& 0.711& 0.750 & 4.36\\
\ml{1}{torch-ash\,\cite{dong2023fast}} & GT & 0.042& 0.056& 0.751& 0.678& 0.710 &0.47\\
\ml{1}{Ours} & GT & 0.042& \textbf{0.043}& 0.776& \textbf{0.748}& \textbf{0.762} & \textbf{0.03}\\
\midrule
\ml{1}{DROID-SLAM\,\cite{teed2021droid}} & SLAM & 0.082& 0.153& 0.504& 0.469& 0.475 & \textbf{0.02}\\
\ml{1}{Ours} & SLAM & \textbf{0.059}& \textbf{0.059}& \textbf{0.663}& \textbf{0.638}& \textbf{0.650} & 0.03\\
\bottomrule
\vspace{-8mm}
\end{tabular}}
\end{table}

\subsection{Evaluation on ScanNet dataset}
We conduct further experiments on ScanNet~\cite{dai2017scannet} to verify our system with real-world datasets, which are notably more challenging due to their larger size and blurry images. We first evaluate the tracking performance by reporting the absolute trajectory error (ATE) metric. To ensure global consistency, both GO-SLAM~\cite{zhang2023goslam} and DROID-SLAM~\cite{teed2021droid} deploy expensive full BA to correct pose drift, whereas our can run the proposed $Sim(3)$-based pose graph BA efficiently in online manner. Tab.~\ref{tab:scannet} shows that our approach achieves on-par accuracy to \revadd{the} strong baseline\revdel{s} \revadd{DROID-SLAM} which employs \revadd{offline} full BA and even some RGB-D methods.

For reconstruction quality, both Tab.~\ref{tab:scannet} and Fig.~\ref{quality-scene0000} show our system yields more accurate and more complete reconstructions than DROID-SLAM. Given the challenging nature of the ScanNet dataset, previous methods \revadd{\cite{yu2022monosdf, guo2022neural, dong2023fast}} have relied on GT poses and offline pipelines to circumvent the problem. For comparison, we run our proposed framework with the GT poses fixed in BA. Tab.~\ref{tab:scannet} shows that our online system not only matches the accuracy of other offline methods but also achieves higher completeness.

\subsection{Result on multi-room apartment scene}
To test our approach quality on even larger scenes, we conduct experiments on the larger-scale apartment dataset~\cite{zhu2022nice} which has over 10k frames and traverses over multiple rooms. We calculate the ATE using the trajectory estimation by the offline RGB-D method~\cite{choi2015robust} as reference. As a result, DROID-SLAM~\cite{teed2021droid} fails to find all loop closures as the drift accumulates very fast and results in collapsed reconstruction (Fig.~\ref{quality-scene0000}), whereas our approach can instantly close the loops once detected reaching a globally consistent map.

\subsection{Ablation Study}
\textbf{Monocular priors.}
First, we investigate the effect of incorporating monocular priors into our system. Tab.~\ref{tab:ablate} shows that without either the normal or depth prior leads to degraded results. The normal prior loss plays a crucial role in improving both accuracy and completeness, whereas the depth prior primarily contributes to accuracy. We further assess the depth L1 metric in Tab.~\ref{tab:ablate}. Even after scale alignment, the monocular prior depth remains suboptimal due to its ill-posed nature. This is also evidenced by the experiment labeled 'w/o BA depth', which we replace the BA depth with the aligned prior depth during mapping. Nevertheless, the prior depth effectively boosts the BA depth estimation when combined with the proposed JDSA module. Finally, our neural map can render further improved depth with an averaged L1 error of 3.63\,cm. 

Furthermore, incorporating depth prior has a positive impact on tracking accuracy. As shown in Tab.~\ref{tab:scannet}, our frontend\,(VO) achieves slightly better ATE than DROID-SLAM\,(VO). Arguably, this can be attributed to the improved depth estimation by the JDSA module, allowing the optical flow network to converge to more correspondences in later iterations. 

\vspace{-3mm}
\setlength{\tabcolsep}{3pt}
\begin{table}[h]
\centering
\caption{Impact of different components based on reconstruction metrics (left) and depth L1 metric (right). Numbers are averaged over 8 sequences of Replica dataset.}\label{tab:ablate}
\vspace*{-3mm}
\resizebox{0.49\textwidth}{!}{
\begin{tabular}{cccc}
\toprule
& Acc[cm]$\downarrow$ & Comp[cm]$\downarrow$ & Comp. Ratio[\%]$\uparrow$\\
\midrule
\ml{1}{w/o BA depth} & 8.23 & 7.83& 64.32 \\
\ml{1}{w/o $\mathcal{L}_{normal}$} & 4.63 & 4.75& 76.80 \\
\ml{1}{w/o depth prior} & 4.23 & 3.98& 81.10 \\
\ml{1}{Ours} & \textbf{3.56} & \textbf{3.60} & \textbf{82.95}\\
\bottomrule
\end{tabular}
\quad

\begin{tabular}{cc}
\toprule
Depth type & L1[cm]$\downarrow$\\
\midrule
Aligned prior & 9.36 \\
BA only & 6.11 \\
BA+JDSA & 4.81 \\
Rendered & \textbf{3.63}\\
\bottomrule
\vspace{-3mm}
\end{tabular}}
\end{table}

\begin{figure}[t!]
\centering
\includegraphics[width=0.49\textwidth]{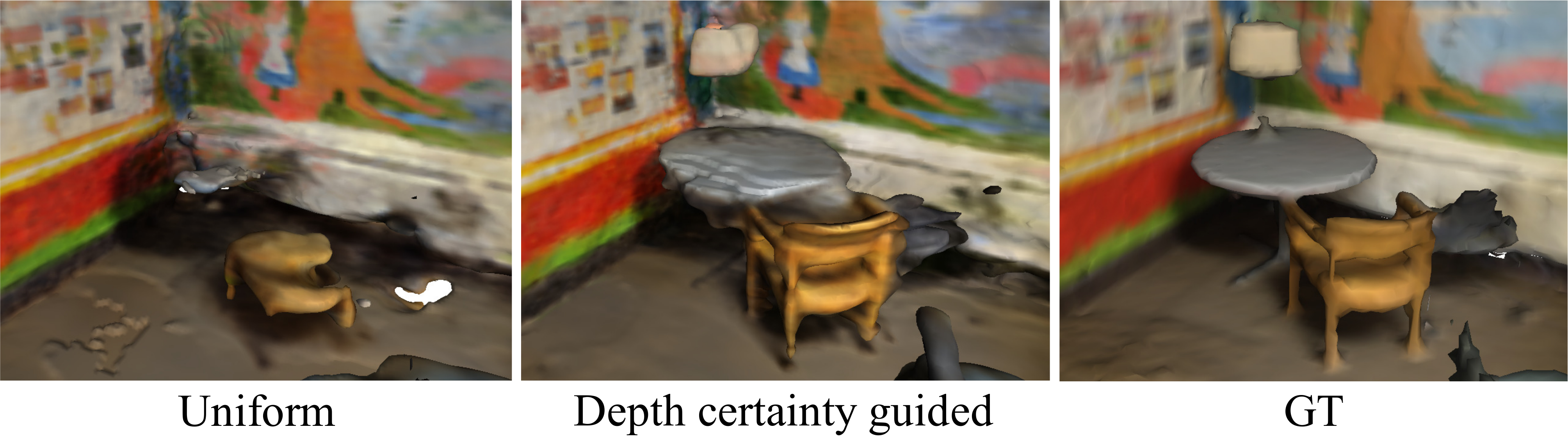}
\vspace*{-5mm}
\caption{Ablation study on uniform sampling vs. depth certainty guided pixel sampling.}
\label{ablate-sampling}
\end{figure}
    
\textbf{Depth guided sampling.}
Fig.~\ref{ablate-sampling} presents the reconstruction results using different pixel sampling methods. The uniform pixel sampling method as employed by prior works~\cite{zhang2023goslam,zhu2023nicer} faces challenges to accurately reconstruct object boundaries and small objects. One potential cause is the high ratio of noisy depth estimations which can falsely carve out the surfaces, leading to conflicts between the SDF loss and free-space loss during map optimization. In contrast, as shown in Fig.~\ref{fig:depth}, we observe that the pixels on small objects and object edges typically attain higher confidences from the optical flow network and corresponding higher certainties. Utilizing the depth certainty-guided pixel sampling method allows for more pixel sampling in these regions and helps the neural fields in distinguishing the contradictory supervisions between the SDF loss and free-space loss.

\begin{figure}[t!]
\centering
\includegraphics[width=0.45\textwidth]{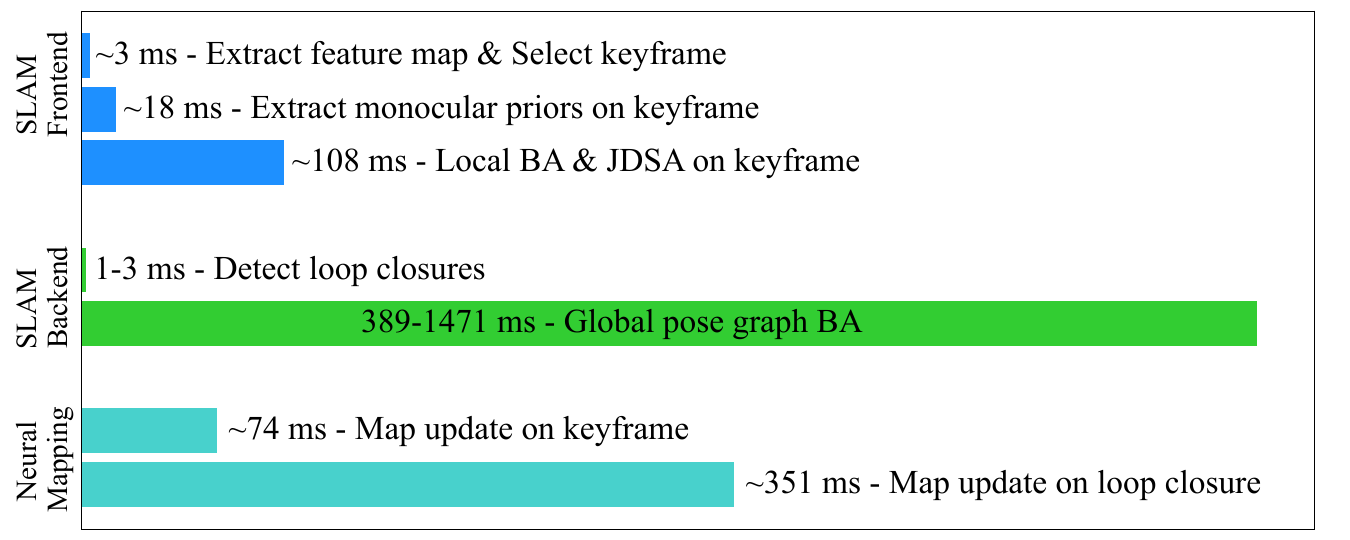}
\vspace*{-3mm}
\caption{Runtime analysis of the key components in each process.}
\label{fig:runtime}
\vspace*{-3mm}
\end{figure}

\vspace*{-3mm}

\subsection{Runtime Analysis}
All experiments are carried out on a desktop PC with an Intel i9 CPU and an Nvidia RTX 4090 GPU. We report the runtime of key components of the system. Fig.~\ref{fig:runtime} shows that majority time is consumed to process keyframes. The frontend can handle up to 5 keyframes per second. The mapping process can rapidly update with camera movement, requiring only 74\,ms for 10 optimization iterations. Notably the loop closing and global optimization run in parallel, taking a few hundreds milliseconds up to around 1.5 seconds. On Replica, our system operates at an average speed of 25 fps. On ScanNet, the speed reduces to 15 fps due to the faster motion and more keyframes need to be processed.

\vspace*{-3mm}
\section{CONCLUSIONS}
In this letter, we present our integration of deep learning-based dense SLAM with neural field representation to reconstruct high-quality scene geometry in real-time. We jointly adjust depth maps and the scales of monocular priors to not only solve the scales of the priors but only enable accurate depth estimation. The estimated depth aids efficient ray sampling and optimization of the neural fields. Consequently, the neural map can be incrementally and continuously constructed in a live manner. To maintain global consistency, our system employs the proposed $Sim(3)$ based pose graph BA when loop closures are detected, correcting both pose and scale drifts. We show that our neural map can instantly adapt to these global updates by loop closures. Compared to previous methods, our approach achieves the state-of-the-art accuracy and completeness. 

\ifCLASSOPTIONcaptionsoff
  \newpage
\fi

\bibliographystyle{IEEEtran}
\bibliography{IEEEabrv,root}








\end{document}